\documentclass[11pt]{article}

\usepackage[preprint]{acl}
\usepackage{colortbl}
\usepackage{times}
\usepackage{latexsym}
\usepackage{booktabs}
\usepackage{subcaption}
\usepackage{enumitem}
\usepackage[T1]{fontenc}

\usepackage[utf8]{inputenc}

\usepackage{multirow}
\usepackage{microtype}
\usepackage{amsmath}
\usepackage{inconsolata}
\usepackage{makecell}
\usepackage{graphicx}

\title{
The Role of Mixed-Language Documents for Multilingual Large Language Model Pretraining
}

\author{Jiandong Shao,$^{2}$ Raphael Tang,$^{1}$ Crystina Zhang,$^{3}$  Karin Sevegnani,$^{4}$ 
\\\textbf{Pontus Stenetorp,$^{1,5}$ Jianfei Yang,$^{2}$ Yao Lu$^{1}$} \vspace{1mm}\\
$^1$University College London~~$^2$Nanyang Technological University\\$^3$University of Waterloo~~~$^4$NVIDIA~~~$^5$National Institute of Informatics
}

\begin{document}
\maketitle
\begin{abstract}

Multilingual large language models achieve impressive cross-lingual performance despite largely monolingual pretraining. While bilingual data in pretraining corpora is widely believed to enable these abilities, details of its contributions remain unclear. We investigate this question by pretraining models from scratch under controlled conditions, comparing the standard web corpus with a monolingual-only version that removes all multilingual documents.
Despite constituting only 2\% of the corpus, removing bilingual data causes translation performance to drop 56\% in BLEU, while behaviour on cross-lingual QA and general reasoning tasks remains stable, with training curves largely overlapping the baseline.
To understand this asymmetry, we categorize bilingual data into parallel~(14\%), code-switching~(72\%), and miscellaneous documents~(14\%) based on the semantic relevance of content in different languages. We then conduct granular ablations by reintroducing parallel or code-switching data into the monolingual-only corpus. Our experiments reveal that parallel data almost fully restores translation performance~(91\% of the unfiltered baseline), whereas code-switching contributes minimally. Other cross-lingual tasks remain largely unaffected by either type. These findings reveal that translation critically depends on systematic token-level alignments from parallel data, whereas cross-lingual understanding and reasoning appear to be achievable even without bilingual data.
\end{abstract}
\section{Introduction}
\begin{figure}[t]
\centering
\includegraphics[width=\columnwidth]{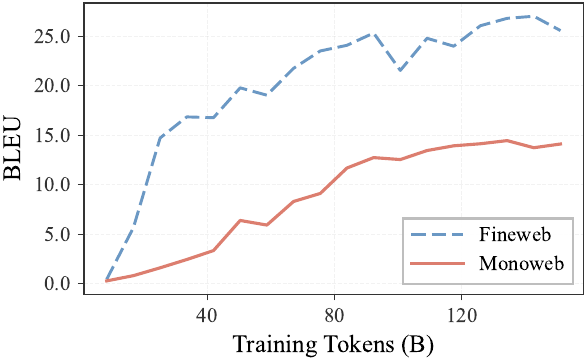}
\caption{
Performance on WMT14 for different pretraining setups. \textsc{FineWeb}: multilingual web data from FineWeb and FineWeb-2; \textsc{MonoWeb}: multilingual web data with bilingual documents \textit{removed}.
}
\label{fig:wmt-enfr}
\end{figure}
Large language models (LLMs), when pretrained on web-collected data from different language sources, exhibit remarkable emergent capabilities in cross-lingual understanding despite not being pretrained using multilingual-specific objectives~\citep{mBERT,Achiam2023GPT4TR,Yang2024Qwen25TR}. 
Existing research attributes this behaviour not only to a sufficient amount of data from different languages, but also to specific documents where multiple languages co-occur in the same context~\citep{Chaudhary2020DICTMLMIM,Chi2020InfoXLMAI,Wang2025InvestigatingAS}. 
Motivated by this observation, multilingual pretraining strategies have often incorporated multilingual data, under the hypothesis that mixed-language exposure uniformly benefits cross-lingual tasks~\citep{Yoo2024CodeSwitchingCL,Wang2025InvestigatingAS}.

\begin{table*}[t]
\centering
\small
\begin{tabular}{p{2.8cm}p{12.5cm}}
\toprule
\textbf{Category} & \textbf{Example} \\
\midrule
\textbf{Parallel} & 
\textcolor{blue}{Magnifique et lumineux loft Toronto de 1 chambre avec plafonds de 10 pi et grande terrasse extérieure comprenant un barbecue....} \\
&Beautiful, bright one bedroom Toronto loft with 10ft ceilings and large outdoor terrace including barbeque...... \\
& \textit{[Paragraph-aligned translation with systematic cross-lingual correspondence]} \\
\midrule
\textbf{Code-switching} & 
The people, filled with joy, chant the anthem \textcolor{blue}{``A qua ben fé! A qua ben fé! La tarascou a rou un bré!''} \\
& \textit{[Natural language mixing within shared discourse context]} \\
\midrule
\textbf{Miscellaneous} & 
...and in some cases whether to let the fires burn to create regeneration in the forest. \textcolor{blue}{Vous devez avoir la dernière version de Flash Player installée.} \\
& \textit{[French text about Flash Player, semantically unrelated to the English content]} \\
\bottomrule
\end{tabular}
\caption{Examples of \textsc{MonoWeb} filtered data. We classify documents into three categories: documents with clear parallel structure, documents that exhibit code-switch behaviour, and miscellaneous documents. 
}
\label{tab:bilingual-examples}
\end{table*}

However, the high cost of pretraining and large-scale pretraining data classfication  has constrained the scope of existing explorations of the role of multilingual data. Studies that rely on continual pretraining typically build on models that may have already been exposed to related data during pretraining, which makes the role of multilingual data more difficult to disentangle.
Among the few works that investigate multilingual data at the pretraining stage, existing studies~\cite{cherry,Qorib2025JustGP,Wang2025InvestigatingAS} do not provide a systematic analysis of its role, but instead focus on specific settings or mechanisms.
%
To this end, we aim to conduct a thorough analysis of the pretraining corpus and design a controlled pretraining setup to explicitly reveal the role of multilingual data.

We construct a monolingual web corpus by filtering out all documents containing more than one language from standard web-collected data. This procedure removes fewer than 2\% of documents, making fine-grained analysis of multilingual data feasible at scale.
We then pretrain multilingual LLMs from scratch under two setups: \textsc{MonoWeb}, using the filtered corpus, and \textsc{FineWeb}, using the original web data.
Despite accounting for only 2\% of pretraining data, multilingual documents are critical for machine translation. Removing them causes BLEU scores to drop from 22.3 to 9.8 (a 56\% relative decrease), effectively collapsing translation performance. In contrast, other cross-lingual tasks are substantially less affected: cross-lingual QA drops by 10\% on average, while understanding and reasoning tasks vary by at most 4\%. This asymmetry highlights the nuanced role of pretraining data across different multilingual tasks.

To better understand this phenomenon, we analyze the composition of the removed multilingual documents. We find that most consist of bilingual content, which can be grouped into three categories (\autoref{tab:bilingual-examples}): (i) \emph{parallel documents} (14\%), providing aligned translations with explicit cross-lingual correspondence, such as multilingual Airbnb webpages; (ii) \emph{code-switching documents} (72\%), where languages naturally alternate within shared discourse, as commonly observed in user-generated content on platforms like Pinterest; and (iii) \emph{miscellaneous documents} (14\%), where multiple languages co-occur without meaningful semantic alignment.
We then isolate the contributions of different bilingual data types through controlled pretraining from scratch. Our results show that parallel data, despite comprising only 14\% of bilingual documents, is the dominant factor for translation performance: reintroducing only parallel data yields a 106\% improvement over \textsc{MonoWeb}, largely recovering performance relative to \textsc{FineWeb} (BLEU $20.2$ vs.\ $22.3$). In contrast, reintroducing code-switching data provides only marginal gains (BLEU $12.4$ vs.\ $9.8$), with little effect on other cross-lingual tasks.
Finally, we analyze the underlying causes of this asymmetry. We find that removing bilingual data primarily disrupts lexical-level cross-lingual alignment, leading to severe translation failures, while sentence-level alignment remains largely preserved, explaining the robustness of non-translation tasks.

To summarise, our contributions are threefold:
\begin{enumerate}
\item We introduce a monolingual dataset, \textsc{MonoWeb}, together with a detailed analysis of multilingual content, pretrain models from scratch to study multilingual behavior without mixed-language exposure, and open-source both the dataset and models.
\item Through pretraining from scratch, we demonstrate a task-dependent sensitivity to bilingual data: machine translation critically depends on a tiny fraction (less than 2\%) of bilingual documents, whereas other cross-lingual understanding and reasoning tasks remain largely unaffected. We further show that different types of bilingual data contribute unequally, with parallel data playing a disproportionately critical role.
\item We provide an in-depth failure mode and representation-level analysis, revealing that the degradation in translation performance is driven by the loss of lexical-level alignment, while sentence-level alignment remains largely preserved.
\end{enumerate}

\section{Related Work}

Multilingual data is widely assumed to drive cross-lingual capabilities in multilingual models. Parallel corpora (sentence-aligned translations) are well-known to be essential for machine translation~\citep{Brown1993TheMO}, enabling multilingual MT systems to bridge high- and low-resource languages~\citep{Johnson2016GooglesMN,Fan2020BeyondEM,team2022NoLL}. Beyond parallel data, naturally occurring code-switching, where languages alternate within the same discourse, has also attracted attention as a potential mechanism for cross-lingual alignment. Prior work shows that using code-switching for data augmentation can improve zero-shot transfer during finetuning~\citep{Qin2020CoSDAMLMC}, and that curriculum learning with code-switching enhances transfer to low-resource languages~\citep{Yoo2024CodeSwitchingCL}. These results have motivated practitioners to incorporate multilingual content under the assumption that mixed-language exposure benefits cross-lingual tasks~\citep{Qorib2025JustGP}.

However, most existing studies focus on finetuning or continued training, which remains limited because models may have already been exposed to similar data during the pretraining stage. Among the few works that investigate multilingual data in the pretraining stage, most focus on specific aspects rather than systematically studying its role: one primarily characterizes incidental bilingualism in existing corpora~\citep{cherry}, another uses generated data to study curriculum learning effects~\citep{Qorib2025JustGP}, and a third explores synthetic code-switching for cross-lingual transfer~\citep{Wang2025InvestigatingAS}. This leaves a gap in the understanding of the role of multilingual data during pretraining, particularly regarding the differential contributions of parallel versus code-switching data. We aim to fill this gap by systematically ablating different bilingual data types and studying their impact on multilingual LLMs.

\section{MonoWeb Pretraining Data}

To study the heterogeneous role of bilingual data, we first construct a multilingual corpus by sampling 60B tokens per language from FineWeb-Edu~\citep{lozhkov2024fineweb-edu} (English) and FineWeb2~\citep{fineweb2} (German, Spanish, French), totaling 240B tokens. We then perform a systematic characterization of bilingual documents in this corpus, focusing on English-paired bilingual content (en-de, en-es, en-fr) as English serves as the current dominant lingua franca for cross-lingual scenarios.

\subsection{Bilingual Data Identification}

We identify bilingual documents through a two-stage pipeline, combining rule-based filtering with LLM-based classification to ensure both scalability and accuracy.

\paragraph{Stage 1: Candidate Detection via Entropy-based Filtering.}
We first detect candidate bilingual documents using language-level entropy as a proxy for language mixing. For each document, we perform sentence segmentation using NLTK~\citep{nltk} and apply fastText language identification~\citep{fasttext} to compute language confidence scores for each sentence. Taking English-French as an example, for each sentence $s_i$ with length $l_i$, fastText outputs confidence scores for English ($p_i^{\text{en}}$) and French ($p_i^{\text{fr}}$). We then compute a document-level language distribution by aggregating sentence-level scores weighted by sentence length:\

\begin{equation}
P_{\text{doc}}^{\text{lang}} = \frac{\sum_{i} l_i \cdot p_i^{\text{lang}}}{\sum_{i} l_i},
\end{equation}
where $\text{lang} \in \{\text{en}, \text{fr}\}$. After normalization, we obtain a probability distribution over the two languages for the entire document. We compute the entropy of this distribution:

\begin{equation}
H = -\sum_{\text{lang}} P_{\text{doc}}^{\text{lang}} \log P_{\text{doc}}^{\text{lang}}.
\end{equation}
Documents with entropy above a threshold $\tau=0.1$ (indicating substantial mixing of both languages) are marked as bilingual candidates.
We empirically selected this threshold by examining the distribution of entropy values and verifying that it effectively captures documents with substantial language mixing. 
This stage serves as a coarse filtering that optimizes for the recall of potential bilingual data, 
while maintaining computational efficiency. 
As a result, 5\% of the corpus is retained and can be precessed more computationally expensive methods during the subsequent verification stage.

\paragraph{Stage 2: LLM-based Classification.}

To distinguish different types of bilingual relationships, we employ \textsc{Llama-3.3-70B-Instruct}~\citep{Llama3} for a \textit{two-step classification} process, whose reliability has been validated through human evaluation. First, the model verifies whether each candidate is genuinely bilingual, which aims to filter out the false negatives  
included by entropy-based filtering.
We consider the resulting set of documents after this step as the final verified bilingual documents, which consists of approximately 2\% of the entire corpus.
Second, based on the semantic relationship of contents in different languages, the verified bilingual documents are classified into one of the three categories:\
\begin{itemize}[leftmargin=*,noitemsep,topsep=2pt]
    \item \textbf{Parallel documents}: Paragraph-aligned translations where languages express identical semantic content with systematic correspondences (e.g., dictionaries, translated website; the example in \autoref{tab:bilingual-examples} is from Airbnb).
    \item \textbf{Code-switching documents}: Documents where both languages appear with semantic relationships but without systematic alignment. This includes naturally occurring mixed-language discourse (e.g., multilingual forum discussions), articles with embedded foreign quotations or terminology, and documents where languages serve complementary communicative functions. Crucially, unlike parallel data, the two languages do not provide translations of each other but rather contribute distinct yet related semantic content.
    \item \textbf{Miscellaneous documents}: Documents where multiple languages co-occur without meaningful cross-lingual semantic relationships. This category primarily consists of web artifacts such as multilingual boilerplate, advertisements in different languages, or navigation elements appended to otherwise monolingual content.
\end{itemize}

This two-stage approach balances computational efficiency with classification accuracy: entropy-based filtering reduces the search space from the full corpus to ~5\% candidates, while LLM classification provides semantic nuance that rule-based methods lack. Table~\ref{tab:bilingual-examples} shows representative examples for each category. The resulting taxonomy enables granular ablations to isolate the effects of different bilingual data types during pretraining.

\begin{table}[t]
\centering
\begin{tabular}{lrrr}
\toprule
\textbf{Data Type} & \textbf{en-de} & \textbf{en-es} & \textbf{en-fr} \\
\midrule
\multicolumn{4}{l}{\cellcolor{gray!15}\textit{Bilingual data in Corpus}} \\
\quad Total Bilingual & 2.80\% & 1.62\% & 2.40\% \\
\midrule
\multicolumn{4}{l}{\cellcolor{gray!15}\textit{Bilingual Data Composition}} \\
\quad Parallel & 10\% & 17\% & 15\% \\
\quad Code-switching & 75\% & 69\% & 73\% \\
\quad Miscellaneous & 15\% & 14\% & 12\% \\
\bottomrule
\end{tabular}
\caption{Bilingual data statistics for each language pair. The top section reports the proportion of bilingual data in the full corpus, showing that such data is generally sparse; the bottom section shows the distribution of the bilingual data types.}
\label{tab:data-composition}
\end{table}

\begin{table}[t]
\small
\centering
\begin{tabular}{lp{4.5cm}c}
\toprule
\textbf{Type} & \textbf{Representative Sources} & \textbf{\%} \\
\midrule
\multirow{3}{*}{Parallel}
& Academic (thesis.fr) & 35 \\
& Dictionaries (reverso.net) & 15 \\
& Travel (airbnb.com) & 15 \\
& Canadian (umontreal.ca) & 6 \\
& Professional (docs.microsoft) & 8 \\

\midrule
\multirow{3}{*}{Code-switching}
& Social (pinterest.com) & 25 \\
& Forums (forumactif.com) & 10 \\
& E-commerce (amazon.fr) & 8 \\
\bottomrule
\end{tabular}
\caption{Approximate domain distribution of bilingual data based on URL analysis of the top 50 sources from the en-fr corpora. Parallel data originates mainly from academic sources and dictionaries with systematic alignments, while code-switching appears in user-generated content with organic language mixing.}
\label{tab:bilingual-sources}
\end{table}

\subsection{Bilingual Data Analysis}

\autoref{tab:data-composition} presents the statistics of bilingual data in our corpus, where the pattern is similar for all three language pairs:\ the bilingual data constitutes 2\% of the entire 240B-token pretraining corpus, and it is dominated by code-switching data ($>$ 70\%), and a similar amount of parallel and miscellaneous documents (10-20\%).
We further analyze the website URLs of the parallel and code-switching documents to understand the main sources of each categories of bilingual data, reported in \autoref{tab:bilingual-sources}.
\smallskip


Parallel data, while comprising less than 20\% of bilingual data, originates from high-quality curated sources.
As \autoref{tab:bilingual-sources} reveals, academic repositories dominate, particularly doctoral theses with multilingual abstracts. Bilingual dictionaries and language learning platforms (reverso.net) provide sentence-aligned translations, while technical documentation (docs.microsoft.com) contributes systematic correspondences. These sources feature explicit token-level alignments where each segment has an equivalent in another language.

Code-switching dominates at 72\% of bilingual data, which originates primarily from social content aggregation sites (pinterest.com), E-commerce with mixed-language reviews (amazon.fr), and forums. 

The remaining miscellaneous 14\% consists of noise—multilingual boilerplate and web artifacts where languages accidentally co-occur without meaningful relationships. 

Overall, the URL analysis reveals a fundamental distinction on the source of bilingual data under different categories:\ parallel data provides professionally curated alignments from dictionaries and academic repositories, while code-switching reflects spontaneous language mixing in user-generated content.

\section{Experimental Setup}

\subsection{Pretraining Configurations}
We conduct experiments on three language pairs: English-French (en-fr), English-German (en-de), and English-Spanish (en-es). For each language pair, we construct a bilingual corpus by combining 60B English tokens with 60B tokens from the target language (French, German, or Spanish), sampled from FineWeb-Edu and FineWeb2.

For each language pair, we pretrain models using four data configurations:

\begin{itemize}
    \item \textbf{\textsc{FineWeb}}: Full 120B-token corpus including all bilingual data.
    \item \textbf{\textsc{MonoWeb}}: All bilingual documents removed, retaining only monolingual content.
    \item \textbf{\textsc{MonoWeb+Parallel}}: \textsc{MonoWeb} augmented with only parallel documents.
    \item \textbf{\textsc{MonoWeb+CodeSwitch}}: \textsc{MonoWeb} augmented with only code-switching documents.
\end{itemize}

We exclude miscellaneous data as it lacks cross-lingual semantic relationships. This yields 12 models in total (3 language pairs × 4 configurations), all trained from scratch to ensure a fair comparison.

\subsection{Model Architecture and Training}

We train decoder-only transformer models with 1.35B parameters using the Llama-2 tokenizer~\citep{Touvron2023Llama2O} (32K vocabulary). The architecture consists of 24 layers with a 2048 hidden dimensions, 16 attention heads, and a 2048 context length. All models are trained for 34K steps (\~143B tokens) with a batch size of 2,048 using the AdamW optimizer~\citep{Loshchilov2017DecoupledWD} and a 6e-4 learning rate, including 2,000 warmup steps followed by constant decay. We set weight decay to 0.1, apply gradient clipping at 1.0, and use Adam betas of 0.9 and 0.95. Training is performed with Megatron-LM~\citep{megatron-lm} and takes about 6,144 A100 GPU hours per model.

\subsection{Downstream Evaluation Suite}
All tasks are evaluated using the lm-evaluation-harness~\citep{eval-harness}, along with five-shot prompting and default configurations.

\paragraph{Machine Translation}
We evaluate translation quality on standard benchmarks for all three language pairs: wmt16 en-de~\citep{WMT16}, wmt14 en-fr~\citep{wmt14}, and flores-101 en-es~\citep{FLORES101}, testing both translation directions and report BLEU scores~\citep{bleu} separately for each direction.

\paragraph{Cross-lingual Question Answering}
We evaluate cross-lingual question answering using two complementary benchmarks. (1) For XQuAD~\citep{XQuAD}, we adapt the dataset by placing the context in language L1 and both the question and answer in language L2, allowing us to assess the model’s ability to generate answers across languages. (2) For MLQA~\citep{lewis2019mlqa}, we follow the original setup, where the context and answer are in language L1 and the question is in language L2, which primarily evaluates the model’s ability to retrieve information across languages. We report Exact Match scores for all language pairs.

\paragraph{Cross-lingual Understanding and Reasoning}
We evaluate models on a suite of benchmarks covering both cross-lingual understanding and reasoning abilities. For cross-lingual natural language understanding, we use XNLI~\citep{XNLI} and PAWS-X~\citep{paws-x} to assess whether bilingual data improves the transfer of inference and paraphrase recognition skills. For reasoning tasks, we include HellaSwag~\citep{Hellswag,Hellswag_arc_multilingual} for commonsense reasoning, ARC~\citep{Hellswag_arc_multilingual,arc} for knowledge-intensive reasoning, TruthfulQA~\citep{truthfulqa,Hellswag_arc_multilingual} for factual consistency, and additionally XStoryCloze~\citep{xstorycloze} (en, es) and XWinograd~\citep{xwinograd} (en, fr) for narrative comprehension and coreference resolution. We report accuracy for all tasks.

\begin{table*}[t]
\small
\centering
\begin{tabular}{p{4.5cm}p{5.2cm}p{5.2cm}}
\toprule
\textbf{Source (en)} & \textbf{\textsc{FineWeb} (de)} & \textbf{\textsc{MonoWeb} (de)} \\
\midrule
The students should receive a grant \underline{immediately}. 
& Die Schüler sollten \textbf{sofort} einen Zuschuss erhalten. 
& Die Studierenden sollten eine Unterstützung erhalten. \\
\midrule
This was a conscious decision \underline{- diversity is an important topic here}.
& Dies war eine bewusste Entscheidung \textbf{- Vielfalt ist ein wichtiges Thema hier}.
& Das war ein bewusster Entschluss. \\
\midrule
He's a hero to his \underline{kids} and his wife.
& Er ist ein Held für seine \textbf{Kinder} und seine Frau.
& Er ist ein Held für seine Familie und seine Frau. \\
\bottomrule
\end{tabular}
\caption{Fine-grained information loss in \textsc{MonoWeb} translations. Core propositions are preserved, but precise details are systematically lost: temporal specifications (example 1: "immediately"), explanatory contexts (example 2: diversity rationale), and lexical precision (example 3: "kids" → "Familie" [family]). Bold text shows precise translations; underlined text indicates lost information.}
\label{tab:information-loss}
\end{table*}

\begin{table}[t]
\centering
\small
\begin{tabular}{lrrrr}
\toprule
\textbf{Direction} & \textsc{FWB} & \textsc{MWB} & \textsc{MWB+P} & \textsc{MWB+CS} \\
\midrule
\rowcolor{gray!5}
en→de & 16.2 & 5.0 & 17.0 & 4.6 \\
de→en & 24.6 & 14.5 & 21.3 & 14.9 \\
\rowcolor{gray!5}
en→es & 17.7 & 6.6 & 17.3 & 11.4 \\
es→en & 21.4 & 8.3 & 20.1 & 16.0 \\
\rowcolor{gray!5}
en→fr & 25.4 & 12.1 & 22.7 & 17.4 \\
fr→en & 28.6 & 15.3 & 28.8 & 18.7 \\
\midrule
\textbf{Average} & \textbf{22.3} & \textbf{9.8} & \textbf{20.2} & \textbf{12.4} \\
\bottomrule
\end{tabular}
\caption{BLEU scores for each translation direction. Removing bilingual data (\textsc{MWB}) causes a substantial drop, while adding parallel data (\textsc{MWB+P}) largely restores performance. FWB = \textsc{FineWeb}, CS = Code-Switch.}
\label{tab:mt-detail}
\end{table}

\section{Results}


Table~\ref{tab:mt-detail},~\ref{tab:qa-results}, and~\ref{tab:understanding-results} present results across all tasks and configurations. A clear task-specific asymmetry emerges: removing bilingual data causes significant degradation for machine translation (56\% BLEU drop), moderate decline in cross-lingual QA (<10\%), and has almost no effect on understanding and reasoning tasks. This indicates different levels of reliance on bilingual exposure across tasks, suggesting that different cross-lingual abilities may rely on qualitatively different learning signals.





\subsection{Machine Translation: Critical Dependence on Parallel Data}

Table~\ref{tab:mt-detail} summarizes translation results across all configurations and language pairs. Removing all bilingual data leads to substantial performance degradation, with average BLEU dropping from 22.3 to 9.8 (56\% relative decline). Reintroducing only parallel documents which comprise 10–17\% of bilingual content, largely recovers performance (20.2 BLEU, 91\% of the original performance). In contrast, adding back code-switched text—72\% of bilingual data—yields a minimal improvement (12.4 BLEU, only 56\% of original performance).

This pattern is consistent across all six translation directions (Table~\ref{tab:mt-detail}). Individual language pairs show a 41–69\% relative degradation when bilingual data is removed, and 90–107\% recovery when parallel data alone is reintroduced. 

These results highlight that translation quality depends critically on explicit cross-lingual alignment rather than incidental code-switching.

\begin{table}[t]
\small
\centering
\begin{tabular}{lcccc}
\toprule
& \textsc{FWB} & \textsc{MWB+P} & \textsc{MWB} & \textsc{MWB+CS} \\
\midrule
German \% & 86.6 & 89.7 & 43.6 & 45.2 \\
DE BLEU & 17.4 & 17.8 & 7.70 & 6.21 \\
\bottomrule
\end{tabular}
\caption{Language generation rate and translation quality on En→De. \textsc{MWB} and \textsc{MWB+CS} fail on both: low German generation (45\% vs. 85\%) and poor quality when generating German (18.5 vs. 25.1 BLEU).}
\label{tab:translation-analysis}
\end{table}

\subsection{Understanding Translation Collapse: A Two-Fold Failure}

To understand the mechanisms behind translation performance degradation, we analyze 1,000 sampled En→De translation outputs, using Llama-3.3-70B-Instruct as a language identifier to classify each as German, English, or mixed-language. Table~\ref{tab:translation-analysis} shows a clear disparity: \textsc{FineWeb} and \textsc{MonoWeb+Parallel} generate German in more than 85\% of cases, while \textsc{MonoWeb} and \textsc{MonoWeb+CodeSwitch} produce German in only around 45\%. The remaining 55\% are predominantly English passthroughs, which naturally yield zero BLEU contribution. These results indicate that models trained without parallel data often fail at the most basic requirement of translation—producing text in the target language.

However, language generation failure alone cannot account for the full extent of BLEU degradation. As shown in Table~\ref{tab:translation-analysis}, when the evaluation is restricted to outputs that are correctly generated in German, \textsc{MonoWeb} still achieves only 7.70 BLEU which is less than half of \textsc{FineWeb}’s 17.4. Even under comparable output-language conditions, a 56\% quality gap remains. \textsc{MonoWeb+CodeSwitch} performs even worse at 6.21 BLEU. This reveals two compounding failure modes: (1) 56\% failure to generate target language (43.6\% vs. 86.6\% German generation), and (2) severely degraded translation quality for the remaining outputs (7.70 vs. 17.4 BLEU). The overall performance decrease compounds both problems.

To further examine the nature of the degraded translation fidelity, we manually analyzed 100 correctly generated German outputs. The analysis reveals a consistent pattern of semantic under-specification: \textsc{MonoWeb} captures only coarse-grained semantics, preserving the basic propositional structure (who does what) but loses fine-grained information about how, when, why, and to what degree. 
As illustrated in Table~\ref{tab:information-loss}, \textsc{FineWeb} accurately preserves temporal and explanatory details ("immediately" → "sofort"; "diversity is an important topic here" → "Vielfalt ist hier ein wichtiges Thema"), whereas \textsc{MonoWeb} tends to produce paraphrases that erase such distinctions. Lexical precision also deteriorates, e.g., translating "kids" as "Familie" [family] instead of "Kinder" [kids].

These observations suggest that without parallel supervision, models internalize only approximate cross-lingual alignment, resulting in content-preserving yet information-thinned translations.

\begin{table}[t]
\centering
\small
\begin{tabular}{l@{\hspace{3pt}}c@{\hspace{3pt}}c@{\hspace{3pt}}c@{\hspace{3pt}}c}
\toprule
\textbf{Task} & \textsc{ FWB } & \textsc{ MWB } & \textsc{ MWB+P } & \textsc{ MWB+CS} \\
\midrule
\rowcolor{gray!15}
\multicolumn{5}{l}{\textit{German}} \\
XQuAD & 28.9 & 25.2 & 31.2 & 29.0 \\
MLQA & 20.6 & 22.4 & 21.4 & 19.1 \\
\midrule
\rowcolor{gray!15}
\multicolumn{5}{l}{\textit{Spanish}} \\
XQuAD & 31.8 & 29.7 & 32.1 & 29.9 \\
MLQA & 22.7 & 23.9 & 22.8 & 20.3 \\
\midrule
\textbf{XQuAD Avg} & \textbf{30.4} & \textbf{27.5} & \textbf{31.7} & \textbf{29.5} \\
\textbf{MLQA Avg} & \textbf{21.7} & \textbf{23.2} & \textbf{22.1} & \textbf{19.7} \\
\bottomrule
\end{tabular}
\caption{Cross-lingual QA performance averaged over both directions per language pair. XQuAD shows moderate sensitivity to bilingual data, while MLQA remains stable. }
\label{tab:qa-results}
\end{table}

\subsection{Other Cross-lingual Tasks: Minimal Dependence on Bilingual Data}


\paragraph{Cross-lingual Question Answering}

\begin{figure}[t]
\centering
\setlength{\abovecaptionskip}{2pt}
\setlength{\belowcaptionskip}{-4pt}

\begin{minipage}[b]{0.225\textwidth}
    \centering
    \includegraphics[width=\textwidth]{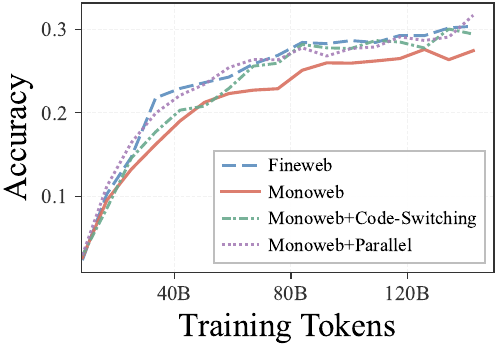}
    \subcaption{}
    \label{fig:xquad-avg}
\end{minipage}
\hfill
\begin{minipage}[b]{0.225\textwidth}
    \centering
    \includegraphics[width=\textwidth]{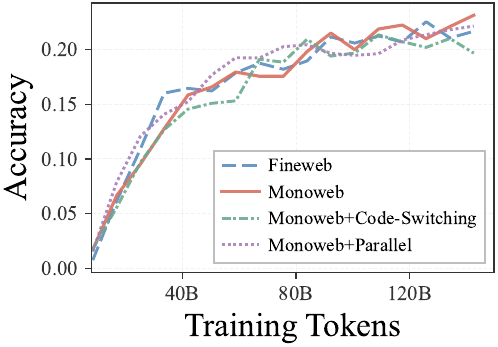}
    \subcaption{}
    \label{fig:mlqa-avg}
\end{minipage}

\vspace{2mm}

\begin{minipage}[b]{0.225\textwidth}
    \centering
    \includegraphics[width=\textwidth]{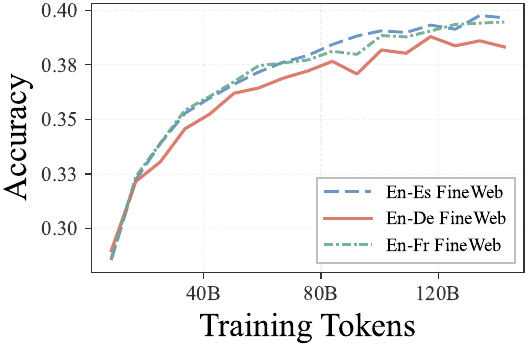}
    \subcaption{}
    \label{fig:hellswag-a}
\end{minipage}
\hfill
\begin{minipage}[b]{0.225\textwidth}
    \centering
    \includegraphics[width=\textwidth]{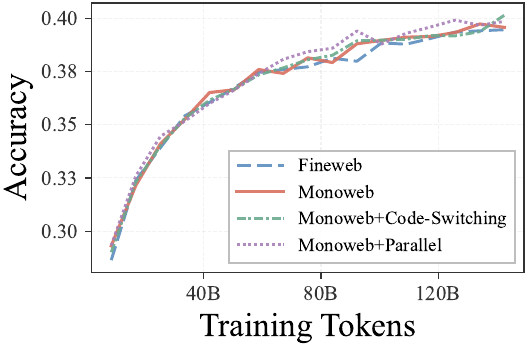}
    \subcaption{}
    \label{fig:hellswag-b}
\end{minipage}

\caption{Training performance across cross-lingual tasks. (a) XQuAD shows consistent separation between configurations, with \textsc{MonoWeb} underperforming throughout training. (b) MLQA exhibits overlapping trajectories across all configurations. (c) HellaSwag performance across language pairs under identical \textsc{FineWeb} setup shows variation, indicating cross-lingual transfer varies by pair. (d) HellaSwag within en-fr: stable performance across bilingual configurations. }

\label{fig:combined-tasks}
\end{figure}

Table~\ref{tab:qa-results} presents results for XQuAD and MLQA. The two tasks show different sensitivity to bilingual data removal.
For XQuAD, \textsc{MonoWeb} underperforms \textsc{FineWeb} throughout training (Figure~\ref{fig:xquad-avg}), achieving 27.5 EM compared to \textsc{FineWeb}'s 30.4 EM (9.5\% drop). On MLQA, the training curves (Figure~\ref{fig:mlqa-avg}) show overlapping trajectories across configurations, with the final scores ranging from 21.7 to 23.2 EM. Unlike XQuAD, MLQA exhibits no consistent separation between configurations during training.
This difference may reflect distinct task structures: XQuAD requires generating L2 answers from L1 contexts, while MLQA primarily involves retrieving answers within L1 after understanding L2 questions. 

\paragraph{Understanding and Reasoning Tasks}
Table~\ref{tab:understanding-results} presents results across five understanding and reasoning benchmarks. All tasks show stability across all bilingual configurations, with performance consistently being within 1-2\% of the baseline.

To better understand this stability, we take HellaSwag as a representative case. Figure~\ref{fig:hellswag-a}, and ~\ref{fig:hellswag-b} demonstrates two complementary findings: First, Figure~\ref{fig:hellswag-a} compares different language pairs under the \textsc{FineWeb} setting where all three pairs use identical English data for 1:1 balanced training, and shows discernible variation in HellaSwag\_En performance, indicating that cross-lingual transfer effects exist and vary across language pairs.
Second, Figure~\ref{fig:hellswag-b} compares different bilingual configurations for the EN–FR pair. Performance remains largely unchanged whether bilingual data is present (\textsc{FineWeb}), absent (\textsc{MonoWeb}), or partially restored (\textsc{MWB+P}, \textsc{MWB+CS}), demonstrating that cross-lingual transfer persists even without bilingual data. Similar patterns also emerge across other benchmarks.

\begin{table}[t]
\centering
\small 
\begin{tabular*}{\linewidth}{@{\extracolsep{\fill}}lcccccc}
\toprule
& \multicolumn{3}{c}{\textbf{Sentence-level P@1}} & \multicolumn{3}{c}{\textbf{Lexical-level P@1}} \\
\cmidrule(lr){2-4} \cmidrule(lr){5-7}
\textbf{Layer} & \textbf{FWB} & \textbf{MWB} & \textbf{$\Delta$} & \textbf{FWB} & \textbf{MWB} & \textbf{$\Delta$} \\
\midrule
0  & 1.7 & 1.8 & +0.2 & 5.8 & 8.3 & +2.5 \\
6  & 60.6 & 55.9 & -4.7 & 40.7 & 19.7 & \textbf{-21.0} \\
12 & 93.7 & 92.5 & -1.3 & 68.7 & 55.1 & -13.6 \\
23 & 81.2 & 79.4 & -1.8 & 25.5 & 18.4 & -7.1 \\
\bottomrule
\end{tabular*}
\caption{\footnotesize Layer-wise Alignment Analysis. Lexical-level alignment shows a sharp drop at middle layers in \textsc{MonoWeb}, while sentence-level alignment remains largely stable.}
\label{tab:alignment_analysis_wide}
\end{table}

\subsection{Explaining the Asymmetry: Why MT Collapses but Reasoning Persists}

Removing bilingual data causes a severe collapse in machine translation, while cross-lingual reasoning and understanding tasks remain largely unaffected. To explain this phenomenon, we analyze how bilingual data removal impacts cross-lingual alignment at different linguistic granularities.
Specifically, we measure alignment across layers using Precision@1 (P@1) computed with cosine similarity for sentence representations (3{,}000 WMT parallel sentences) and word representations (2{,}000 MUSE~\cite{MUSE} pairs).
As shown in \autoref{tab:alignment_analysis_wide}, we observe a stark divergence: while MonoWeb preserves robust sentence-level alignment ($<2\%$ drop from FineWeb), it suffers a sharp 13--21\% degradation in lexical-level alignment.
This suggests that monolingual pretraining is sufficient to align sentence-level semantics, supporting cross-lingual understanding and reasoning, but fails to establish the fine-grained lexical correspondences required for accurate translation.

\begin{table}[t]
\centering
\small
\begin{tabular}{l@{\hspace{3pt}}c@{\hspace{3pt}}c@{\hspace{3pt}}c@{\hspace{3pt}}c}
\toprule
\textbf{Task} & \textsc{FWB} & \textsc{ MWB} & \textsc{ MWB+P} & \textsc{  MWB+CS} \\
\midrule
\multicolumn{5}{l}{\textit{English (Avg)}} \\
\quad XNLI & 46.3 & 45.6 & 45.9 & 46.8 \\
\quad HellaSwag & 39.1 & 39.4 & 39.6 & 39.6 \\
\quad ARC\_C & 32.3 & 33.6 & 34.7 & 33.8 \\
\quad ARC\_E & 68.5 & 68.3 & 68.3 & 67.9 \\
\quad PAWS & 54.5 & 54.9 & 54.0 & 55.5 \\
\quad TruthfulQA & 22.0 & 21.8 & 22.4 & 20.9 \\
\quad Xwinograd & 75.7 & 75.4 & 74.0 & 73.6 \\
\quad Xstorycloze & 64.6 & 65.2 & 64.1 & 65.6\\

\midrule
\multicolumn{5}{l}{\textit{German}} \\
\quad XNLI & 44.5 & 43.4 & 43.8 & 41.4 \\
\quad HellaSwag & 34.8 & 35.0 & 35.5 & 35.2 \\
\quad ARC & 22.9 & 24.1 & 24.9 & 25.2 \\
\quad PAWS & 51.9 & 52.0 & 51.6 & 51.8 \\
\quad TruthfulQA & 23.4 & 21.4 & 21.3 & 24.1 \\
\midrule
\multicolumn{5}{l}{\textit{Spanish}} \\
\quad XNLI & 43.5 & 42.3 & 43.9 & 45.4 \\
\quad HellaSwag & 38.6 & 38.6 & 38.5 & 39.2 \\
\quad ARC & 28.6 & 29.7 & 27.9 & 28.5 \\
\quad PAWS & 50.1 & 53.1 & 51.3 & 51.8 \\
\quad TruthfulQA & 25.6 & 26.7 & 25.2 & 26.7 \\
\quad Xstorycloze & 62.3 & 61.6 & 61.6 & 61.4 \\
\midrule
\multicolumn{5}{l}{\textit{French}} \\
\quad XNLI & 44.6 & 44.0 & 44.0 & 44.5 \\
\quad HellaSwag & 38.0 & 38.5 & 38.4 & 37.9 \\
\quad ARC & 29.1 & 26.4 & 26.5 & 26.9 \\
\quad PAWS & 52.6 & 47.9 & 52.2 & 53.8 \\
\quad TruthfulQA & 24.4 & 25.8 & 22.9 & 25.4 \\
\quad Xwinograd & 61.5 & 66.3 & 61.5 & 60.2 \\
\bottomrule
\end{tabular}
\caption{
Multilingual understanding and reasoning performance across all language pairs. 
For English, the reported numbers are averaged over several language pairs.
%
}

\label{tab:understanding-results}
\end{table}

\section{Conclusion}
This study explored the role of bilingual data in multilingual LLM pretraining and uncovered a clear task asymmetry. Translation is highly sensitive to a small fraction of bilingual content (2\%), whereas other cross-lingual tasks remain largely unaffected. Further analysis show that parallel data, not code-switching text, drives translation performance. This indicates that explicit cross-lingual alignment is essential for translation, while monolingual exposure largely suffices for broader cross-lingual understanding. These findings imply that multilingual pretraining may benefit more from high-quality parallel data than from large quantities of code-switched text. More broadly, our results highlight that the impact of bilingual data during multilingual pretraining can vary substantially across tasks, suggesting that its role is nuanced even within the pretraining stage.


\section{Limitations}
Our study has several limitations. First, due to computational constraints, we pretrained only 1.35B-parameter models and did not pretrain larger models such as 7B, which may exhibit different sensitivity to bilingual data. Second, our experiments focus on major languages within the Latin script family, leaving open questions about the impact of bilingual data on typologically distant or low-resource languages. Third, our analysis categorizes bilingual data into parallel, code-switching, and miscellaneous types, but finer-grained distinctions, such as domain, register, or sentence-level alignment quality, may further influence cross-lingual learning.

\bibliography{custom}




\end{document}